# Stochastic Sensitivity Analysis Using Fuzzy Influence Diagrams

Pramod Jain, Graduate Research Assistant
Alice M. Agogino, Associate Professor

*Intelligent Systems Research Group, aagogino@kepler.berkeley.edu*
*Dept. of Mechanical Engineering, 5136 Etcheverry Hall*
*University of California, Berkeley, CA 94720*

*Abstract* - The practice of stochastic sensitivity analysis described in the decision analysis literature is a testimonial to the need for considering deviations from precise point estimates of uncertainty. We propose the use of Bayesian fuzzy probabilities within an influence diagram computational scheme for performing sensitivity analysis during the solution of probabilistic inference and decision problems. Unlike other parametric approaches, the proposed scheme does not require resolving the problem for the varying probability point estimates. We claim that the solution to fuzzy influence diagrams provides as much information as the classical point estimate approach plus additional information concerning stochastic sensitivity. An example based on diagnostic decision making in microcomputer assembly is used to illustrate this idea.

## INTRODUCTION AND OBJECTIVE

In discrete probability theory point estimates quantifying the likelihood of uncertain events are represented as crisp numbers. Unfortunately, even for moderately complex random processes, it can be extremely difficult to specify the probabilities of uncertain events to much precision. There is always a "fuzziness" associated with subjective assessments of uncertainty. There are Bayesians who claim that classical probability theory is equipped to handle this fuzziness (e.g., [Cheeseman:1985], [Kyburg: 1987] and [Pearl:1987]) and others who argue the opposite case and propose various schemes for higher order probabilities ([Bonissone: 1985], [Dempster: 1967], [Shafer: 1976] and [Zadeh: 1984]). However, even Bayesians show the need for considering deviations from precise estimates of point probabilities through the accepted practice of stochastic sensitivity analysis ([Howard: 1968] and [Holstein: 1973]). Howard (1968) claims that *stochastic sensitivity is a powerful tool for locating the important variables* and Holstein (1973) is of the view that it *helps in determining the level of encoding of uncertainty and risk attitude*. We propose that fuzzy probabilities used within an influence diagram framework allow the use of linguistic probabilities when communicating with humans concerning uncertainty and an efficient computational scheme to perform a kind of stochastic sensitivity analysis. We do not argue that fuzzy probabilities are second-order probabilities - but that they can be convenient communication and computational tools. We insist, however, in defining Bayesian fuzzy probabilities so that they are consistent with Sage's axioms of probability theory.

Probability theory is one of the oldest and most widely used formalisms for representing and processing uncertain information. Discrete probabilities quantifying an individual's degree of belief of precisely defined events are encoded and Bayes' rule is used to perform probabilistic inferences given the input data. Recently probabilistic influence diagrams (also called Bayes' networks) have emerged as a tool for both graphically representing probabilistic influences and propagating uncertainties through the network ([Shachter: 1985], [Pearl: 1987], [Rege & Agogino:1988], [Heckerman & Horvitz:1987], and [Henrion & Cooley: 1987]). With the addition of decision nodes, influence diagrams also provide a representational scheme and calculus for solving complex decision problems involving uncertain variables ([Miller et al.: 1976], [Howard & Matheson: 1984], [Shachter: 1986], and [Agogino & Rege: 1987]).

In spite of the normative virtues of probability theory [Cheeseman: 1985], descriptive studies of human behavior reveal serious limitations. Cognitive scientists and knowledge engineers have documented the difficulty and biases involved in obtaining point estimates of probabilities from experts ([Kahneman et al.: 1985], [Zimmer: 1983]). Few of us can respond like Mr. Spock in the science fiction television series *Star Trek* and cite probabilities to three decimal places. The linguistic modeling approach, on the other hand, overcomes the requirement of numerical precision. Zimmer (1983,1985) reports that verbal expressions of uncertainty were more accurate than numerical values in estimating the frequency of multiple attributes in his experimental studies. These *fuzzy probabilities* are everyday linguistic expressions of likelihoods, such as *very likely, unlikely*, and *impossible*.

Fuzzy set theory was introduced by Zadeh (1965). It provides a mechanism for representing and manipulating vagueness in practical systems for computer-based decision making ([Zadeh: 1983], [Dubois & Prade: 1980]). Linguistic variables are used to represent imprecise information in a manner similar to natural language and fuzzy operators provide the inference mechanism. There have been several applications of the use of fuzzy linguistic variables in the field of process control [Kickert & Mamdani: 1978] and natural language processing [Zadeh: 1978,1981 and 1975]). Various researcher have developed techniques to map human linguistic expressions of uncertainty to point estimates of probability by means of calibration experiments and fuzzy set theory ([Zimmer: 1985], [Wallsten et al.: 1986], and [Jain & Agogino: 1988a]). The fuzzy set approach, however, is by no means devoid of numerical definitions; it may be viewed as a higher level of complexity beyond conventional point-estimate numerical methods ([Nguyen: 1979], [Tsukamoto & Hatano: 1983]). Further, opponents argue that fuzzy logic is an inadequate approximation to the more desirable calculus of probability theory.

Although there is a rich body of literature associated with probability theory and fuzzy number theory separately, there has been little work done in the area of integrating these seemingly different fields of study. Recently, Jain & Agogino (1988b) have specified the properties of a Bayesian fuzzy probability and have developed arithmetic operations that are consistent with Bayes' rule and retain closure of the required properties. The arithmetic operations developed are those necessary for Bayesian analysis: addition, multiplication, division, and expectation of joint and marginal discrete probability distributions. Application of the arithmetic operations results in a solution in which the mean of the fuzzy function is equivalent to the point estimate obtained by using conventional Bayesian probability. The resulting fuzzy function around the mean can be used for sensitivity analysis; its interpretation depending on the application.

We claim that fuzzy influence diagrams provide efficient computational techniques that contribute to both fuzzy set theory and Bayesian decision analysis. The advantages are two-fold: (1) they allow the use of fuzzy linguistic variables for representing uncertainty while maintaining the desirable properties and rigor of conventional probability theory and (2) they provide a computationally efficient method to perform sensitivity analysis in Bayesian inference and decision analysis.

## BAYESIAN FUZZY PROBABILITIES : BASICS

In this paper, we will assume a discrete random process with precisely defined or nonfuzzy events, the likelihood of which are represented by Bayesian fuzzy probabilities. Bayesian fuzzy probabilities cannot be treated like ordinary fuzzy numbers because they are constrained by the axioms of probability theory:

- The domain of the fuzzy probability is restricted to the closed interval [0,1].
- The sum of all the fuzzy probabilities of collectively exhaustive and mutually exclusive events in a state must be equal to the nonfuzzy number 1.

Definition: A *Bayesian fuzzy probability FP* is a convex normalized fuzzy set $A$ of [0,1] (an interval on the real line $R$), such that

i) there exists a unique $x_0$ (also called the mean) $\in [0,1]$ satisfying $\mu_A(x_0) = 1$,
ii) $\mu_A$ is continuous on the open interval $(0,1)$,
iii) the mean of $FP$ satisfies the axioms of conventional probability theory,
iv) if $\Omega$ is the sample space of events then $FP(\Omega) = 1$,
v) if events $E_1$ and $E_2$ are mutually exclusive, then $FP(E_1 \cup E_2) = FP(E_1) \oplus FP(E_2)$.

In the following sections we will drop the Bayesian qualifier, thus referring to Bayesian fuzzy probability as fuzzy probability. By $FP(E)$ we mean the fuzzy probability of event $E$ and by $\oplus$ we mean the fuzzy probabilistic sum. Notice that fuzzy probability is defined only on the interval $[0,1]$. What we mean by this is that, the membership of any point $z \notin [0,1]$ corresponding to the fuzzy probability is 0. An example of a fuzzy probability could be a linguistic expression with a membership function as shown in Fig. 1 [Jain & Agogino, 1988a].

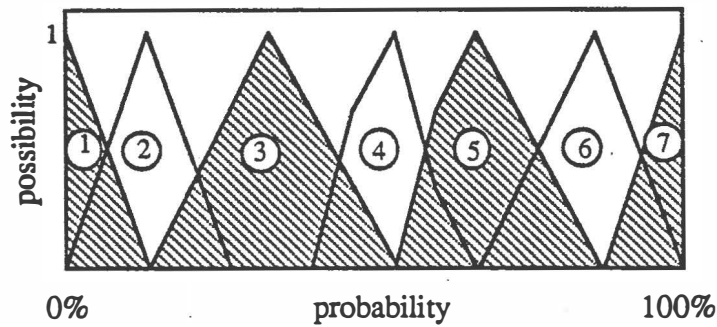

**Fig.1.** Example of fuzzy probabilities: 1. *extremely low*, 2. *quite low*, 3. *low*, 4. *even*, 5. *high*, 6. *very high*, and 7. *extremely high*

Representation of Fuzzy Probabilities. In this section we introduce the $\mathbb{P}$ representation of fuzzy probabilities. We will restrict our attention to only those fuzzy probabilities that have linear membership functions. We will therefore approximate the membership function obtained by performing arithmetic operations on fuzzy probabilities to be linear. The rationale behind this is that a linear approximation is sufficient for the purposes of sensitivity analysis and that without it the problem of manipulating them becomes extremely complex. We make a distinction between three types of fuzzy probabilities:

>*type-0* : zero membership at probability 0 and 1,
>*type-1* : nonzero membership at probability 1 and
>*type-2* : nonzero membership at probability 0.

The $\mathbb{P}$ representation of the three fuzzy probabilities is defined as triplets of the form, $(-(a_m)_{min}, m^*, (a_m)_{max})$ for type-0, $((a_m)_{min}, m^*, [\mu_r])$ for type-1 and $([\mu_s], m^*, (a_m)_{max})$ for type-2, where $a_{min}, a_{max}, \mu_r$ and $\mu_s$ are as shown in Fig. 2; $m^*$ is the mean, $m$ is the base variable and $a_m$ is the spread variable of fuzzy probability $M$. For type-1 and type-2 fuzzy probabilities the term in the square brackets refers to the value of $\mu$ at probabilities 1 and 0, respectively. The base variable and the spread variable are related by the following formula,

$$m = m^* + a_m. \tag{1}$$

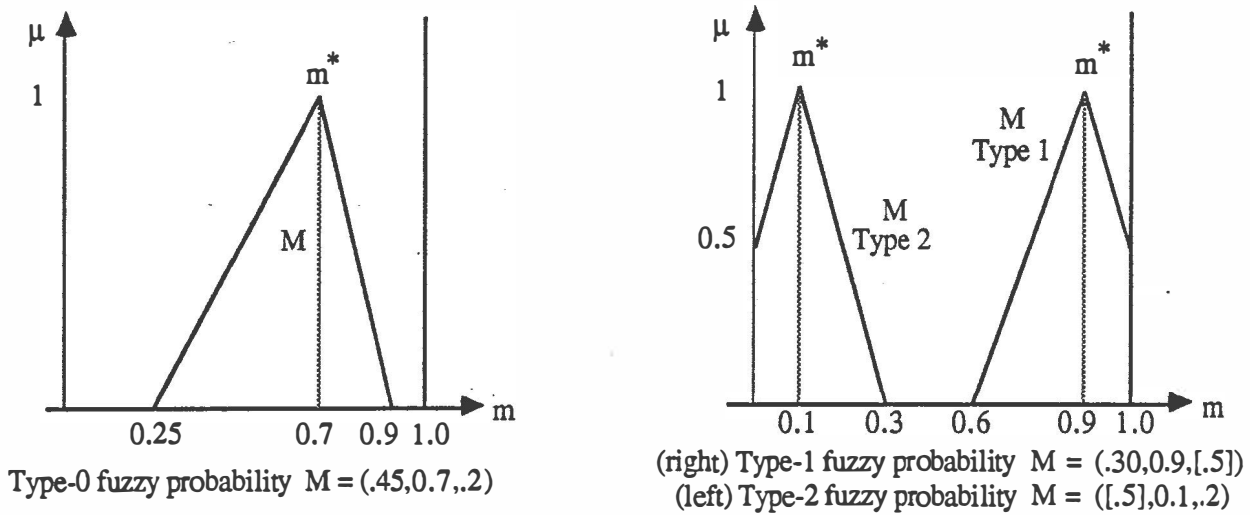

Type-0 fuzzy probability M = (.45,0.7,.2)

(right) Type-1 fuzzy probability M = (.30,0.9,[.5])
(left) Type-2 fuzzy probability M = ([.5],0.1,.2)

**Fig.2.** Illustration of the three types of fuzzy probabilities

<u>Joint and Conditional Fuzzy Probabilities</u> Let $X = \{x_1, ..., x_n\}$ and $Y = \{y_1, ..., y_m\}$ be two universes. If $u_i(X) = FP(X=x_i)$ and $u_j(Y) = FP(Y=y_j)$ are the marginal fuzzy probabilities of events $X = x_i$ and $Y = y_j$ respectively, then let us denote the joint fuzzy probability that $X = x_i$ and $Y = y_j$ by $FP(X=x_i, Y=y_j) = v_{ij}(X,Y)$, and the conditional fuzzy probability of $X = x_i$ given $Y = y_j$ by $FP(X = x_i / Y = y_j) = w_{ij}(X/Y)$. We will drop $X$ and $Y$ when it is obvious which universe we are referring to. Let $u_i^*$, $v_{ij}^*$ and $w_{ij}^*$ be the means and $b_i$, $c_{ij}$ and $d_{ij}$ be the spread variables of the respective probabilities. Since $\Sigma_i u_i = 1$, the following condition must be satisfied by the spread variables $b_i$:

$$\Sigma_i \; b_i (X) = 0. \qquad (2)$$

Similarly the conditional and joint probabilities satisfy the following relationships,

$$\Sigma_i \; w_{ij} (X|Y) = 1, \; \forall j \qquad (3a)$$
$$\Sigma_i \; d_{ij} (X|Y) = 0, \; \forall j \qquad (3b)$$
$$\Sigma_i \; \Sigma_j \; v_{ij} (X,Y) = 1 = \Sigma_j \; \Sigma_i \; v_{ij} (X,Y) \qquad (4a)$$
$$\Sigma_i \; \Sigma_j \; c_{ij} (X,Y) = 0 = \Sigma_j \; \Sigma_i \; c_{ij} (X,Y) \qquad (4b)$$

The joint and the conditional fuzzy probabilities are also related by the following formula, where $\otimes$ is the symbol for fuzzy multiplication.

$$v_{ij} (X,Y) = w_{ij} (X|Y) \otimes u_j (Y) = w_{ij} (Y|X) \otimes u_j (X) \qquad (5)$$

<u>Arithmetic Operations on Fuzzy Probabilities</u>: Since we are dealing with linear membership functions it is sufficient to know the membership of two probabilities, one on either side of the mean, to construct the entire fuzzy probability. Jain & Agogino (1988b) analyze arithmetic operations involved with transformations on type-0, type-1 and type-2 fuzzy probabilities within an influence diagram computational scheme (e.g., IDES: Influence Diagram based Expert System, [Agogino & Rege: 1987]). We will not elaborate on this further except comment on the complexity of the arithmetic operations. The number of ordinary arithmetic operations to perform one fuzzy arithmetic operation is roughly three times. For instance, consider the operation of reversing an arc in an influence diagram - with point probabilities one arc reversal involves $(n-1)$ additions, $n$ multiplications and 2 divisions; with fuzzy probabilities the same operation takes $3(n-1)$ additions, $3n$ multiplications, 6 divisions and $m \cdot log(m)$ comparisons, where $n$ is the number of events in the goal nodes and $m$ is the number of events in the conditioning node. Interval analysis would take roughly twice the number of operations as compared to point probabilities, but it loses information about the mean and gives a larger spread. Notice that the use of fuzzy probabilities obviates the need for evaluating the entire influence diagram at several point probabilities (sweeping the range of a trial value of an aleatory variable) as suggested by Howard (1968). Although it may take approximately 3 times the computation, this is a local effect and can be isolated around the fuzzy variables in an influence diagram solution procedure.

## FUZZY PROBABILISTIC INFERENCE

The solution to an inference problem using probabilistic influence diagrams involves finding conditional probability of a goal node conditioned on a given set of nodes. In a fuzzy probabilistic inference problem the above probabilities are the corresponding fuzzy probabilities. Thus it is equivalent to determining the fuzzy probability distribution FP (G (A) | B) where B is the set of conditioning nodes and G(A) is the goal-value function of the set of nodes A. One procedure for solving an inference problem consists of the following,

- Introduce a goal-value node G(A) with conditioning arcs from nodes of set A.
- Apply transformations sequentially until the desired representation FP(G(A) | B) is obtained, that is, the final influence diagram with arcs leading from the conditioning nodes into the goal node is obtained.

<u>Example of a Fuzzy Inference Problem: The Computer Assembly Diagnostics Problem</u>: As an illustration of the techniques discussed so far let us consider a concrete problem [Agogino & Rege, 1987]. The problem we will examine arises in automated assembly of microcomputers. To simplify the presentation we will assume that after complete assembly, final testing is performed on the microcomputer and a failure in it can be traced to failures in either of two components: (1) the logic board or (2) the I/O board.

Let us further assume that a sensor is available that gives rudimentary information about the operating status of the assembled microcomputer and is a function of the operating status of the logic board and the I/O board. The associated influence diagram is shown in Fig. 3. This represents the Diagnostician's Inference Problem at the topological level.

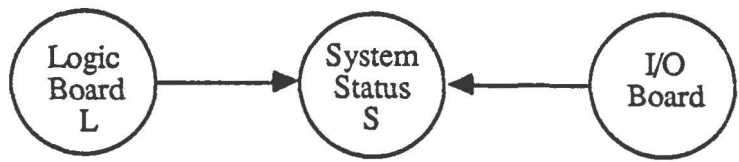

Fig. 3. Probabilistic influence diagram

In the present structure of the influence diagram, the joint fuzzy probability can be obtained from the conditional fuzzy probability of the system status "S" and the fuzzy marginal distributions on the state of the logic board "L" and the I/0 Board "I/0" as given in the expansion below. Note that L and I/O are independent (there is no arc between them) and thus their joint probability distribution is the product of each marginal distribution.

$$FP(S, L, I/O \mid H) = FP(S \mid L, I/O, H) \otimes FP(L \mid H) \otimes FP(I/O \mid H) \qquad (6)$$

The nature of the influences is specified at the functional level. The influence diagram in Fig. 3 implies that the fuzzy conditional distribution on S is known along with the fuzzy marginal distributions on L and I/O. Let us assume for illustration that the test for system status gives a <u>deterministic</u> result based on the status of the I/O and logic boards: if any of these boards (or both) is in a failure state the system status will show a failed system state. Using a subscript of "0" for a failed state and "1" for the operational state, this implies at the numerical level that the conditional distribution of the system status, S is:

$$\Pr(S \mid L, I/O, H) = \begin{cases} 1 & \text{for } S=S_1 \text{ given } L=L_1 \text{ and } I/O=I/O_1 \\ 1 & \text{for } S=S_0 \text{ given } L=L_1 \text{ and } I/O=I/O_0 \\ 1 & \text{for } S=S_0 \text{ given } L=L_0 \text{ and } I/O=I/O_1 \\ 1 & \text{for } S=S_0 \text{ given } L=L_0 \text{ and } I/O=I/O_0 \\ 0 & \text{elsewhere} \end{cases} \qquad (7)$$

Let us further assume that the point estimate of the marginal probability of failure is 5% for the logic board and 1% for the I/O board and that the fuzzy marginal probability for failure is (.03,.05,.03) and ([.66],.01,.03), a deviation of 6% and 4% from the point estimate, respectively.

$$FP(L \mid H) = \begin{cases} (.03, 0.95, .03) & \text{for } L=L_1 \\ (.03, 0.05, .03) & \text{for } L=L_0 \end{cases} \qquad (8)$$

$$FP(I/O \mid H) = \begin{cases} (.03, 0.99, [.66]) & \text{for } I/O=I/O_1 \\ ([.66], 0.01, .03) & \text{for } I/O=I/O_0 \end{cases} \qquad (9)$$

In solving the inference problem, the fuzzy joint probability distribution associated with influence diagram in Fig. 3 is first obtained. The conditional probability FP(I/O | S,H) is then computed by initially integrating over all the states of L and then dividing by FP(S | H) using the arithmetic operations described earlier.

$$FP(I/O \mid S,H) = \begin{cases} ([0.5], 0.1681, .5076) & \text{for } I/O=I/O_0 \text{ given } S=S_0 \\ (.5076, 0.8319, [.5]) & \text{for } I/O=I/O_1 \text{ given } S=S_0 \\ 1 & \text{for } I/O=I/O_1 \text{ given } S=S_1 \\ 0 & \text{elsewhere} \end{cases} \qquad (10)$$

The membership function of $FP(I/O_1 \mid S_0, H)$ is plotted in Fig. 4. It can be clearly seen that the result is very sensitive to the fuzziness around the point estimate of the input probabilities. An advantage of such an analysis over the more common interval analysis, whose output would be a range no smaller than [0.32,1.0], is graphically illustrated in Fig. 4. In our analysis information about how each of these point estimate values are weighted is also provided. In this instance, by observing the distribution of weights associated with the point estimates, one can infer that it is more likely that the probability of the event is close to 1.0 rather than close to 0.32.

183

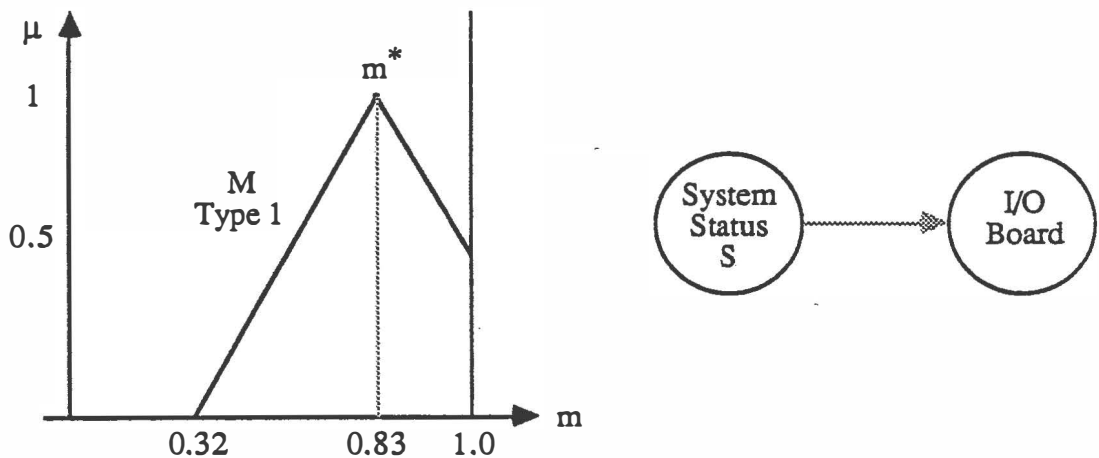

**Fig. 4.** Plot of the membership function of fuzzy probability FP(I/O | S,H) and associated influence diagram

## SOLVING DECISION PROBLEMS

The transformations involved for solving the decision problem, within an influence diagram framework, are the same as for probabilistic inference - arc reversal and node removal. However, instead of determining the likelihood of occurrence of events, the expected value of a sequence of decisions is computed. An optimal sequence of decisions is selected that maximizes the expected value of the value function. Note that arcs to and from decision nodes cannot be reversed. This arises from informational nature of arcs to the decision node (that is, it shows exactly which variables will be known by the decision maker at the time the decision is made) and causal nature of arcs from the decision node.

Example of a Fuzzy Decision Problem: Let us continue with the computer assembly diagnostics example, but now consider the following decision problem - after having computed the likelihood of failure of the two components which component should be tested for failure first and repaired if found defective. In an automated assembly operation, this could mean deciding where the computer should be sent for "rework". If the diagnostician is wrong in his or her decision for rework, valuable time would have been wasted by the rework technician and hence in producing the final product. For purposes of illustration, let us assume that the costs of rework, which involves opening the computer and pulling out the appropriate board, testing, and repairing it, is a constant for each board. Let us define the relevant costs to be minimized as follows:

$$\$Debug_L = \$150 \text{ for the logic board debugging}$$
$$\$Debug_{I/O} = \$100 \text{ for the I/O board debugging}$$
$$\$Repair_L = \$50 \text{ for the logic board repair after debugging}$$
$$\$Repair_{I/O} = \$100 \text{ for the I/O board repair after debugging}$$

The influence diagram in Fig. 5 has been modified to show the decision and value node. It has been assumed that a system failure has occurred and this information is known at the time that the rework decision is made. We will assume for the purposes of this illustration that the debugging and repair costs are known precisely. Situations when costs or risk aversion (utility function parameters) are fuzzy can be handled in a similar fashion.

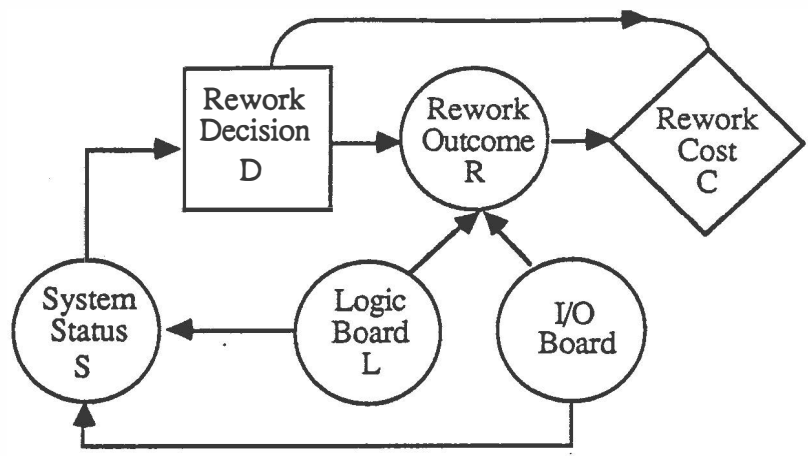

**Fig. 5.** Diagnostician's decision problem

The diagnostician has two choices given that the system status shows a failure ($S=S_0$):
(1) $D_L$ = Send the logic board to rework first or (2) $D_{I/O}$ = Send the I/O board to rework first.

<u>Diagnostician's Decision Problem Solved</u>: Let us now work out the problem at the numerical level, using the probabilistic data given previously. If we assume that the debugging test is "perfect" and influenced only by the states of the logic board, I/O board, and rework decision as shown in Fig. 5, the conditional probability distribution of the results "R" is the trivial deterministic function of the actual states of the logic and I/O boards assuming <u>any</u> initial debug decision is made:

$$\Pr(R \mid L, I/O, D, H) = \begin{cases} 1 \text{ for } R=L_1,I/O_0, \text{ given } L=L_1 \text{ and } I/O=I/O_0 \text{ for either } D \\ 1 \text{ for } R=L_0,I/O_1, \text{ given } L=L_0 \text{ and } I/O=I/O_1 \text{ for either } D \\ 1 \text{ for } R=L_0,I/O_0, \text{ given } L=L_0 \text{ and } I/O=I/O_0 \text{ for either } D \\ 1 \text{ for } R=L_1,I/O_1, \text{ given } L=L_1 \text{ and } I/O=I/O_1 \text{ for either } D \\ 0 \text{ elsewhere} \end{cases}$$

The cost function (value node) depends on the rework outcome "R" and the rework decision "D".

$$C(R, D) = \$\text{Debug}(R, D) + \$\text{Repair}(R) \tag{11}$$

Absorbing the Rework Outcome node "R", the expected value of the cost for a particular decision given the state of the I/O and the logic board is obtained as shown in Fig. 6. Next, the fuzzy expected cost for each of the decisions is computed by doing the following transformations: reverse arc between L and S, absorb node L, reverse arc between S and I/O and finally absorb node I/O.

$$<C(S,D)> = \begin{cases} (26, \$226, 78) & \text{for } S=S_0 \text{ and } D=D_L \\ (50, \$285, 15) & \text{for } S=S_0 \text{ and } D=D_{I/O} \end{cases} \tag{12}$$

Which rework decision minimizes the expected cost ? If one considers the range of possible outcomes in equation (12) above, it is not clear which decision - rework the logic or I/O board - is the best under the circumstances.

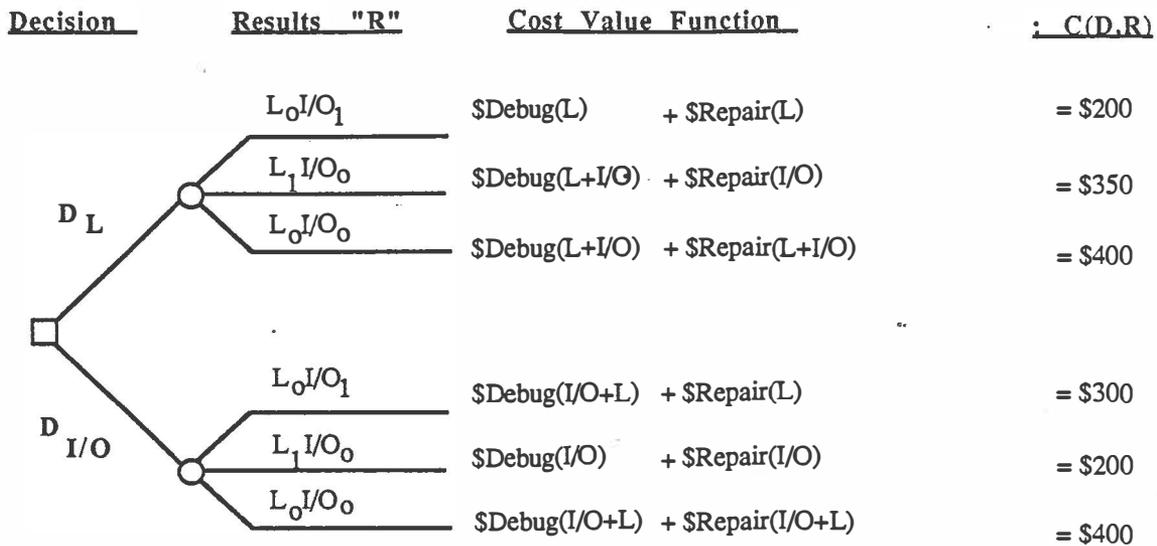

**Fig. 6.** Conditional expected value of C(D,L,I/O) in the form of a decision tree

Criteria for Decision Making: The simplest criterion would be to pick the option with the lowest mean expected cost. However, this would be suboptimal if the point estimate of the input probabilities were not accurately known and were to lie in a region which yields decisions that are not optimal based on the expected cost. We would like to derive some useful information from the membership function of the expected cost that will aid in the process of decision making. We will use the concept of $\alpha$-*intersections* to generate a criterion.

In simple terms we will compute the value of the largest $\alpha$-*intersection* (maximum membership at the intersection of corresponding membership halves) of all the expected costs under consideration, see Fig. 7. Let us call this value $\alpha^*$. It has the following interpretation, with a possibility of one minus $\alpha^*$ or greater the decision based on the point estimate of fuzzy probabilities (same as lowest mean of the expected costs) is optimal unconditionally.

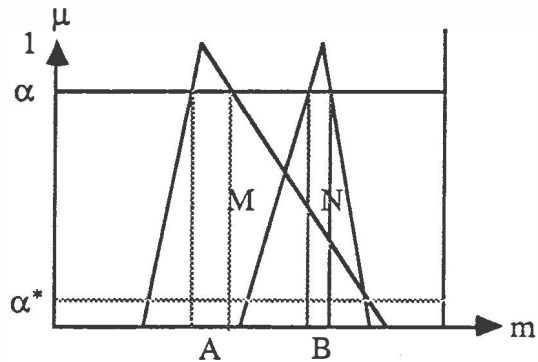

**Fig. 7.** Illustration of $\alpha$-cut sets: A is the $\alpha$-cut of M; B is the $\alpha$-cut of N

$\alpha^*$ can be interpreted as a sensitivity parameter. That is, if $\alpha^*$ is low, the decision is insensitive to input variations in the point estimate. If one decision deterministically dominates the other, then $\alpha^*$ will be zero. If the decision problem can be reduced to one with a single fuzzy probabilistic predecessor of the value node, $\alpha^*$ can also be guaranteed to be zero under first-order stochastic dominance. On the other hand, if $\alpha^*$ is close to one, there is an indication that the expected value of the decision is highly sensitive to the point estimate and that there is a high value for more information or precision associated with the input estimates of probability.

For the general case of multiple variables with fuzzy probabilities, it may be advantageous to consider an alternative to the $\alpha^*$-*intersection* approach. Prior to taking expectation, the difference between the membership functions at each expected cost, subject to constraints that require feasibility and consistency of the probabilities, could be compared. Dominance would be implied if the possibility of the difference membership function is either strictly positive or strictly negative.

Example: Let us continue with the computer assembly diagnostics problem. The membership functions of expected costs in equation (12) are shown in Fig. 7 (M represents $<C(S_0,D_L)>$ and N represents $<C(S_0,D_{I/O})>$). The two right halves of the membership functions intersect at $\alpha^* = 0.064$. Hence the possibility that the decision made based on the point estimates is unconditionally optimal is high, around 94%. This indicates that the decision is relatively insensitive to the value of the point estimate of probability, within the range considered here.

## CONCLUSIONS

An approach to using fuzzy probabilities at the numerical phase of the solution to an inference or decision problem using influence diagrams has been proposed. The use of fuzzy probabilities in influence diagram based expert systems was motivated by the issues that arise in knowledge engineering - reluctance of experts in initially stating precise numerical probabilities and in communicating uncertain information to the user of such a system. In addition, use of fuzzy probabilities provides an efficient computational scheme for obtaining information about the sensitivity of the solution to changes in the point estimates of the probabilities of events.

## ACKNOWLEDGEMENTS

The authors wish to acknowledge the advice and discussions with Ashutosh Rege, who initiated the current research into fuzzy influence diagrams [Agogino & Rege: 1987]. This work was funded, in part, by the National Science Foundation Grant DMC-8451622.